\documentclass{article}

\usepackage{arxiv}

\usepackage[utf8]{inputenc} % allow utf-8 input
\usepackage[T1]{fontenc}    % use 8-bit T1 fonts
\usepackage{hyperref}       % hyperlinks
\usepackage{url}            % simple URL typesetting
\usepackage{booktabs}       % professional-quality tables
\usepackage{amsfonts}       % blackboard math symbols
\usepackage{nicefrac}       % compact symbols for 1/2, etc.
\usepackage{microtype}      % microtypography
\usepackage{lipsum}		% Can be removed after putting your text content
\usepackage{graphicx}
\usepackage{natbib}
\usepackage{doi}

\title{Patent Sentiment Analysis to Highlight Patent Paragraphs}

\author{
 Renukswamy Chikkamath \\
 University of Passau\\
  Passau, Germany\\
   \And
 Vishvapalsinhji Ramsinh Parmar \\
  University of Passau\\
  Passau, Germany\\
  \And
 Christoph Hewel \\
  BETTEN \& RESCH \\
  Munich, Germany\\
  \And
 Markus Endres \\
  University of Passau\\
  Passau, Germany\\
}

\date{}

\renewcommand{\headeright}{Highlight Patent Paragraphs}
\renewcommand{\undertitle}{A Dataset}
\renewcommand{\shorttitle}{}

\hypersetup{
pdftitle={A template for the arxiv style},
pdfsubject={q-bio.NC, q-bio.QM},
pdfauthor={David S.~Hippocampus, Elias D.~Striatum},
pdfkeywords={First keyword, Second keyword, More},
}

\begin{document}
\maketitle

\begin{abstract}
Given a patent document, identifying distinct semantic annotations is an interesting research aspect. Text annotation helps the patent practitioners such as examiners and patent attorneys to quickly identify the key arguments of any invention, successively providing a timely marking of a patent text. In the process of manual patent analysis, to attain better readability, recognising the semantic information by marking paragraphs is in practice.  This semantic annotation process is laborious and time-consuming. To alleviate such a problem, we proposed a novel dataset to train Machine Learning algorithms to automate the highlighting process. The contributions of this work are: i) we developed a multi-class, novel dataset of size 150k samples by traversing USPTO patents over a decade, ii) articulated statistics and distributions of data using imperative exploratory data analysis, iii) baseline Machine Learning models are developed to utilize the dataset to address patent paragraph highlighting task, iv) dataset and codes relating to this task are open-sourced through a dedicated GIT web page: \url{https://github.com/Renuk9390/Patent_Sentiment_Analysis}, and v) future path to extend this work using Deep Learning and domain specific pre-trained language models to develop a tool to highlight is provided.  This work assist patent practitioners in highlighting semantic information automatically and aid to create a sustainable and efficient patent analysis using the aptitude of Machine Learning. 
\end{abstract}

% keywords can be removed

\section{Introduction}
\label{sec:introduction}
\begin{figure}[ht]
  \centering
  \includegraphics[width=9cm]{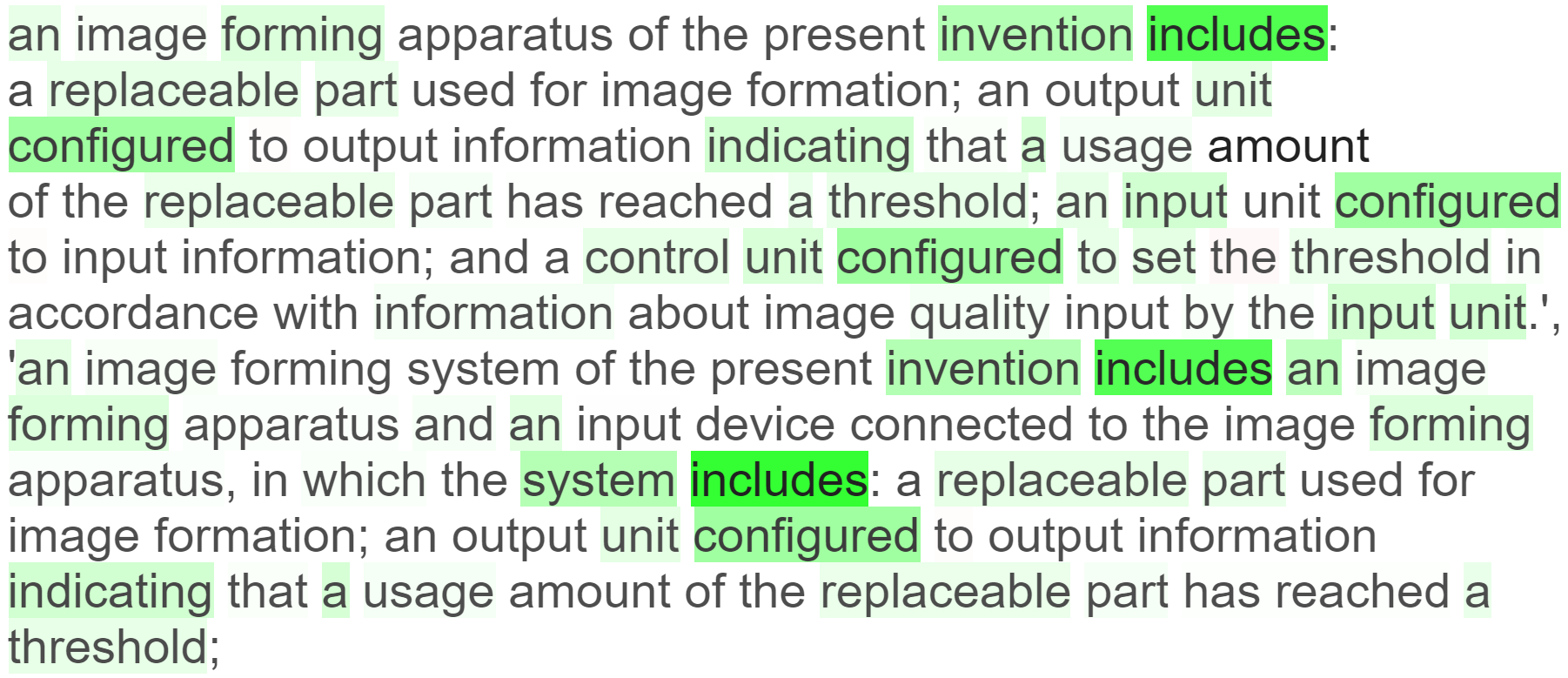}
  \caption{Highlighting important text respect to appropriate title}
  \label{fig:high_text}
  %\Description{Highlighting important text respect to appropriate title}
\end{figure}

Patents are the authority awarded monopolies, granted for innovations that are novel, inventive, and non-obvious in nature. Any individual who wishes to get a grant for an idea must draft a detailed technical patent application document. Further application documents are filed at patent offices, undergo an extensive examination process at respective patent offices. This is where the role of an examiner comes into the picture. Oftentimes, examiners need a careful reading of patent applications in order to find the relevant prior art to the technical subject field. The prior art searching can be done in patent applications, other patent databases, and any non-patent literature. Oftentimes because of indexing in databases, a quick listing of documents can be done. Although there is a list of prior art documents, this is still a laborious and time-consuming process for examiners to mark relevant technical subject matters in time and make a decision on the inventive step of the application. Marking or highlighting such subject matters are crucial activities, which are manually done in practice. Oftentimes patent examiners and attorneys highlight the text passages which could be essential entities for key arguments when patent documents are compared to the existing literature. The important entities such as ``technical advantages'' of any invention, ``problems'' associated with previous efforts, or other plain descriptive texts aka ``boilerplate'' text are manually highlighted by attorneys to compare and contrast individual subject matters. The highlighted text not only helps the examiner to write a detailed search report but also aid in formal hearings to overcome if any objections from patent applicants. With the help of trained machine, text in the paragraph can be highlighted as shown in  Figure \ref{fig:high_text}.

On the other hand, patent applicants or their representatives such as attorneys also read various patent documents in support of their client. In such cases, there is a necessity to keep track of various advantages of the invention in comparison to the prior art. Because this defines the scope of the claims, as the scope of claims becomes broader then the examination becomes more complex and critical. It may end up with several iterations of amendments\footnote{\url{https://www.epo.org/law-practice/legal-texts/guidelines.html}} as per the reports from examiners. Patent attorneys also make use of such manual marking of paragraphs of relevant semantic text. This gives us a shred of evidence that patent paragraph highlighting is one of the important sub-task in patent analysis. The rest of the paper is organized as follows: Section \ref{sec:motivation} provides motivation to perform this task and also its scope towards the patent community. Section \ref{sec:related_wrok} articulates other related work in this direction to highlight text. Section \ref{sec:dataset} discloses the method to generate a dataset and also a detailed exploratory analysis on the same. Section \ref{sec:outcomes} describes the models and their performances in highlighting aka patent sentiment analysis process. Section \ref{sec:conclusion} concludes the paper with future works where a convergence of Deep Learning (DL) models to this end are exposed.

\section{Motivation and Scope}
\label{sec:motivation}
Patent application drafting varies around the globe in accordance with region, and also specific to literature styles. For instance, the innovations or applications drafted from the Asian region often specify the key matters of the invention in specific headers or sections in a patent. These technical matters are often effective technical advantageous of the invention, in other cases, they try to mention selling points of the invention with a separate heading such as “Advantageous effects of invention (AEI)”. Similarly, there are various other annotations that are possible in differentiating technical subject matters such as “Technical problems (TP)” associated with previous efforts or any other boilerplate (plain descriptive text, solutions to problem (SP)). Such annotations or dedicated sections help to wade through the patents and also increase the readability by keeping track of pieces of evidence. However, such pre-defined annotations in patents are not common in all patent documents.

In recent years, the flooding effect\footnote{\url{https://www.wipo.int/edocs/pubdocs/en/wipo_pub_941_2020.pdf}} of patent applications has greatly increased the workload at examination offices to prosecute inventions. To effectively engage in such prosecution, the examiner has to perform a wide range of activities such as a search for prior art, evaluation of inventiveness within the boundaries of patent law, and also to provide a critical assessment of decision in a form report or hearing. In the recent past, few approaches \cite{chikkamath20,krestel21} shown interest in aligning patent analysis strategies using the scope of DL methodologies to address complex patent processes.  In such a complex process, simple highlighting automation can ease the documentation and mark of relevant paragraphs for judgment for discerning the non-obviousness of inventions. To this end, we propose a dataset to train a patent paragraph highlighting models, by which patent practitioners work on patent documents in a much-assisted way and keep track of their evidence. To the best of our knowledge, there is no evidence in the literature that focuses on dataset and training models for patent paragraphs highlighting based on Machine Learning (ML). The dataset can be used to train algorithms on both sentence-level and paragraph levels.

\section{Related Work}
\label{sec:related_wrok}
Patent annotation is a form of information retrieval, conducted in a variety of settings in literature such as keywords or rule-based mining  \cite{li07,rodriguez15,iwayama03}, supervised learning \cite{nanba08,guangpu11} to extract data, and other semi-supervised approaches \cite{brin98, agichtein00} too. However, it is observed that there are no interests shown in the recent past towards highlighting patent paragraphs using ML. Chen \cite{chen15} proposed to use annotation on retrieving effect classes from patent abstract, and further extend the dataset just with few labeled samples. There are associated drawbacks in state-of-the-art such as manual moderations to keyword extraction, semi-automated methods to increase the supervised data generation with labels for training. Unlike our method, the literature mentioned above is purely based on information extraction based on syntax and the semantic nature of the text. However, we propose to use the contextual nature of the text, both on sentence-level and paragraph-level to identify the important subject matters in patent text. In addition, other pre-existing markup-based and rule-based approaches proposed to annotate patent documents \cite{agatonovic08}. This helped to an extent in identifying metadata of patent. However, such rule-based and markup-based methods face difficulties in identifying the arguable subject matters and contextual features of patents. 

Other patent literature based annotations include semantic annotations based on ontology \cite{wang14}. Such an Ontology-based approach again depends on a manually built initial set of patterns. Ontology-based methods pose sub-optimal results in identifying contextual features. Some of the paid solutions such as PATSEER\footnote{\url{https://patseer.com/2021/02/accelerate-your-patent-reviewing-with-patseers-multi-color-highlighting/}} and Patsnap\footnote{\url{https://help.patsnap.com/hc/en-us/articles/115005478629-What-Can-I-Do-When-I-View-A-Patent-}} proposed to highlight patent text automatically, but this is only possible when users know the initial set of keywords and solutions are not open-sourced. As the name indicate, they focus only on highlighting keywords, however in patent literature vocabulary is very rich and diverse. Oftentimes, patent applicants come up with new terminologies, therefore keyword-based approaches often focused less on atomizing patent analysis tasks. A recent interest showed by private IP professionals to use Artificial Intelligence in highlighting patent text by IPGoggles\footnote{\url{https://ipgoggles.com/}}, however, the AI algorithms proposed to use are trained on general English literature, and also no evidence of whether the tools will be made public. To the best of our knowledge, there are no open-sourced datasets and trained models for patent paragraphs highlighting as an annotation process. We propose a novel dataset to highlight patent paragraphs automatically, again this is a multi-class labeled dataset where a pertinent evidence for the correctness of gold standard labels is given in Section \ref{sec:dataset}.

\section{Dataset}
\label{sec:dataset}
\subsection{Data Collection}
\begin{figure*}[ht]
  \centering
  \includegraphics[width=15cm]{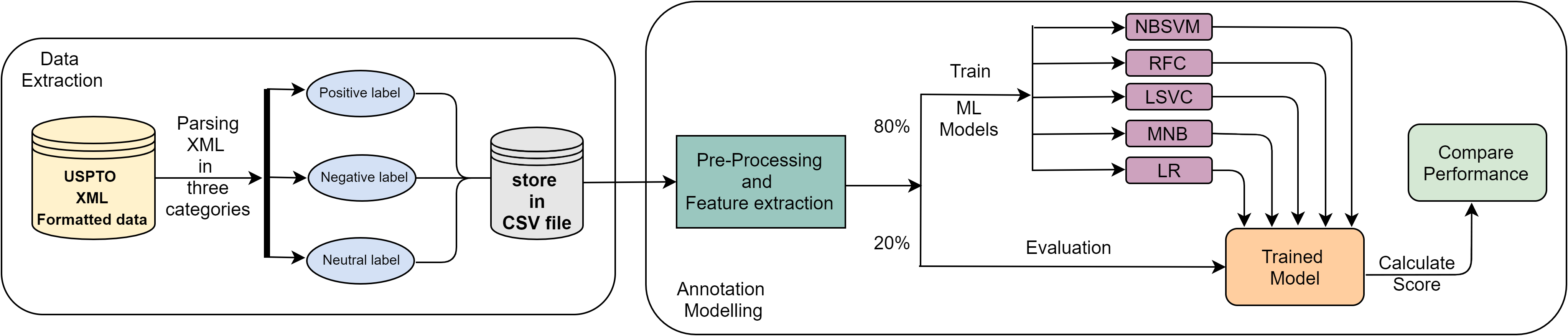}
  \caption{XML file parsing and modelling}
  \label{fig:xml_parsing}
  %\Description{XML parsing for pos, neg, and neutral dataset}
\end{figure*}

Patent sentiment analysis dataset is a curated, selectively extracted collection from the United States Patent and Trademark Office (USPTO) raw XML files. USPTO, to drive advances in innovations and creativity,  provides patent grants\footnote{\url{https://developer.uspto.gov/product/patent-grant-full-text-dataxml}} with full text in nested XML formatted files to the public. For every year, there are grants published are stacked weekly in zipped files (for eg: ipg200107.zip, first week of January 2020). For instance, the link\footnote{\url{https://bulkdata.uspto.gov/data/patent/grant/redbook/fulltext/2020/}} contains 52 XML files, which are arranged according to every week of the year 2020. Each of 52 XML files is again nested structures, contains all published grants for that particular week. We parsed XML files and collected data in a CSV file, the general workflow of this work is shown in Figure \ref{fig:xml_parsing}.

% \begin{figure}[h]
  
%   \centering
%   \includegraphics[width=\linewidth]{acmart-primary/samples/images/XML_parsing.pdf}
%   \caption{XML file parsing for label generation}
%   \label{fig:xml_parsing}
%   \Description{XML parsing for pos, neg, and neutral dataset}
% \end{figure}

%\todo{VP; Add general workflow of the work, modify image from your expose (remove DL, pre-trained models, etc..) and add here, images as small as possible and as clear as possible}

Based on our investigation in Google patents\footnote{A Japan based patent with explicitly mentioned tags (AEI, TP, and SP): \url{https://patents.google.com/patent/US10834907B2/en?oq=US10834907B2}} and interviews with a domain expert\footnote{\url{https://www.bettenpat.com/en-team-hewel.html}}, we found that patent drafting practices and form varies across the globe. For instance, the patent applications originating from Asian countries preferred to explicitly mention several selling points of their invention and drawbacks associated with their referred sources in order to increase the chances of patenting. Such clearly defined structures as dedicated sections in patents help the examiners and attorneys to wade through patent quickly in order to identify key highlights and arguable subject matters. Therefore, we decided to look for such special tags which could be crucial entities during the examination and comparing any existing literature. As shown in Figure \ref{fig:xml_parsing}, three possible tags are selected, however, these tags are uncommon in all patents. In other words, oftentimes availability of such tags is scattered, for example, for the year 2020, out of nearly 200000 grants only around 8000 patents were found with such explicit text segments. So, it is necessary to generate a dataset with special text segments and train algorithms further to automatically highlight the text in any patents where there are no special tags to ease patent analysis. 

\begin{figure}[ht]
  \centering
  \includegraphics[width=12cm]{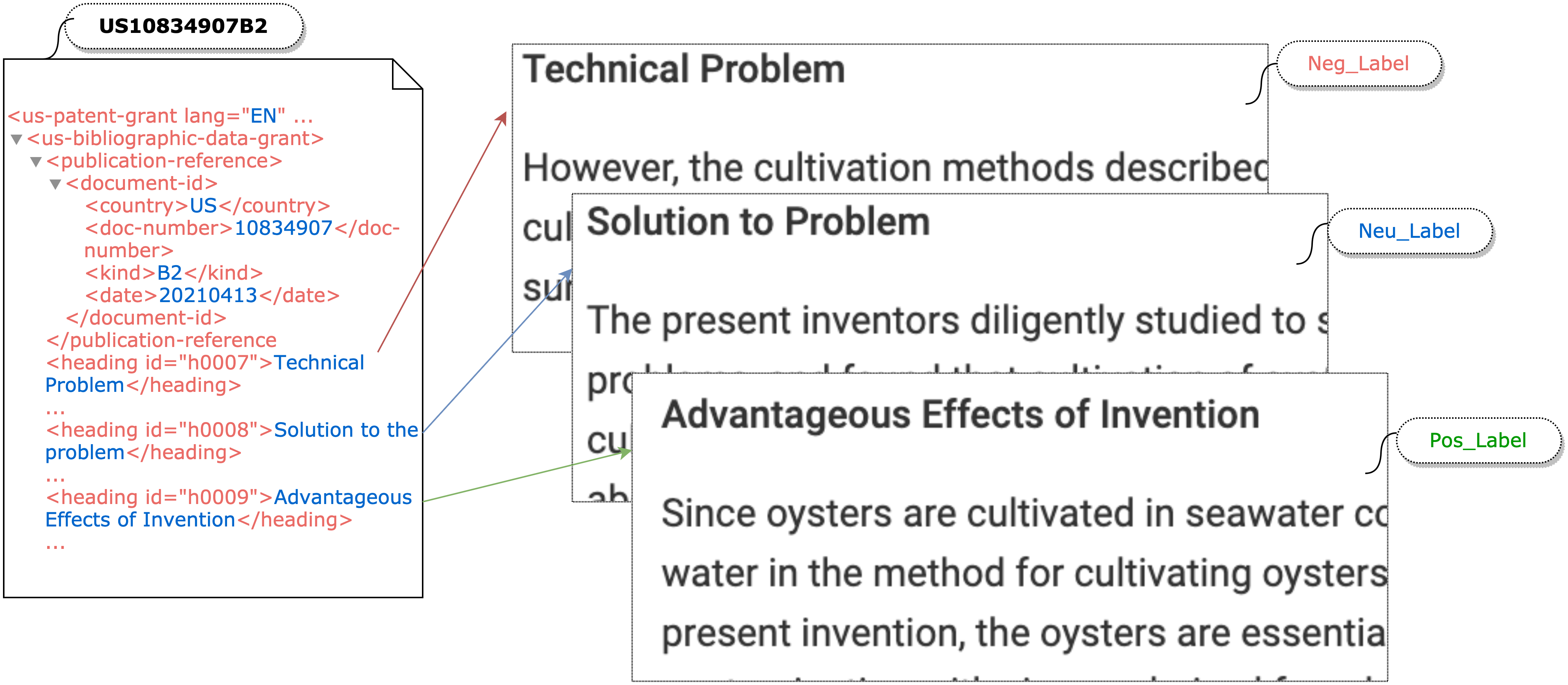}
  \caption{Label attachments according to text descriptions}
  \label{fig:text_labels}
  %\Description{Label attachments according to text descriptions}
\end{figure}

As shown in Figure \ref{fig:text_labels}, “interested data” are considered as three different classes in our dataset. The text paragraphs following the mentioned tags are the text segments (a list of paragraphs) for training any ML algorithms. Three classes namely positive, negative, and neutral vary with their samples every year, however, to have the class balance we retain only an equal number of samples in each class. Along with text, other information such as title, and publication number is collected and made available in the dataset. A sample format of the data with respective labels is shown in Table \ref{tab:dataset}.

\begin{table}[ht]
\centering
  \caption{Sample dataset}
  \label{tab:dataset}
  \begin{tabular}{cccl}
    \toprule
   Doc Num&  Title& Text &  Label\\
    \midrule
    US10842211B2 &  Heat-retaining article&  With the clothing.. &  2 \\
    US10757923B2 &  Aquaculture system&  An embodiment of.. &  0 \\
     US9855011B2 &  Measurement device&  According to the.. &  1\\
     ... &  ...&  ... &  ...\\
  \bottomrule
\end{tabular}
\end{table}

% width=13cm, height=5.5cm
% \begin{figure*}[h] 
% \end{figure*}

%\todo{VP: add image or screenshot of how our dataset looks like in csv maybe, full textwidth, horizontal, and with three samples with labels}

\subsection{Exploratory Data Analysis}

\begin{figure}[ht]
  \centering
  \includegraphics[width=6.15cm, height=7.75cm]{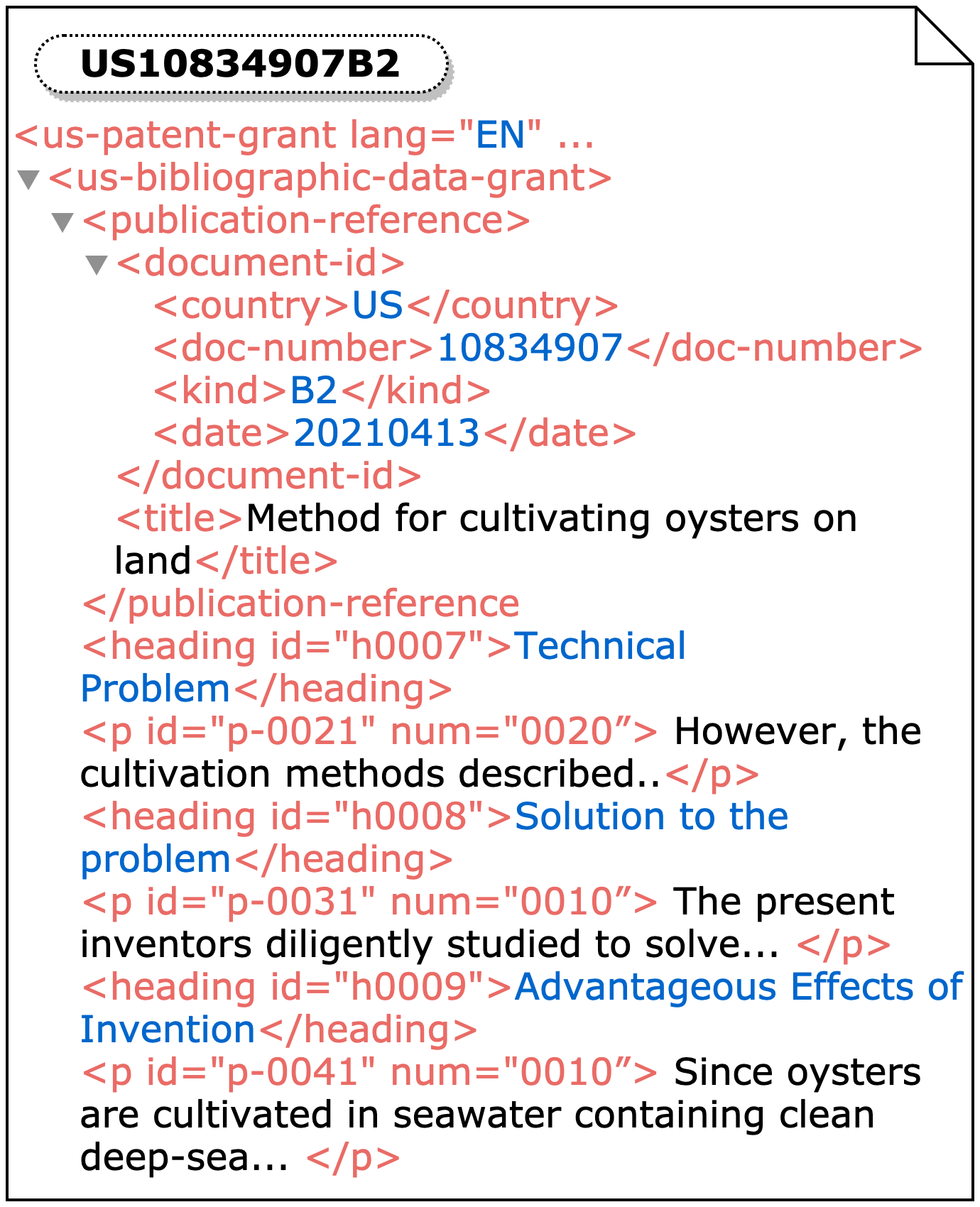}
  \caption{A sample XML patent file}
  \label{fig:xml}
  %\Description{Relation between word counts and paragraph}
\end{figure}

The XML parsers we developed in this work identifies the interested tags (AIE, TP, and SP) and retrieve the paragraphs relating to particular tags. The text content is organised in various paragraphs, each paragraph is identified with unique number in the XML document, for instance <p id="p-0021" num="0020”> as shown in Figure \ref{fig:xml}, similarly different paragraphs are collected in a list. 
The number of paragraphs and words that occur in each paragraph often varies and this again depends on the applicant's drafting style and scope of the invention. The total number of paragraphs in positive samples ranges from 0-41, for negative 0-25, and for neutral samples 0-140, some of the paragraphs count for each category is mentioned in Table \ref{tab:par_counts}. Number of paragraphs (\# Par) and total number available in dataset (Count) are recorded for each class. For instance, maximum number of paragraphs available in  “Neg” class is 25 and respective total count in the dataset is 2. Similarly for other classess, where null value indicated non-availablities of data, for instance maximum paragraphs for “Pos” class is 41 however, there is null value for tuple with 25, means to say there are no patents having positive samples with a size of 25 paragraphs in the dataset. 

\begin{table}[ht]
\centering
  \caption{Total number of paragraphs  }
  \label{tab:par_counts}
  \begin{tabular}{cccl}
    \toprule
   \textbf{(\# Par)}& & \textbf{(Count)} &\\
   &  Pos & Neg & Neu\\
    \midrule
    0 &  27&  73 &  75 \\
    1 &  39562&  23194 &  10401 \\
    .. &  ..&  .. &  .. \\
    \midrule
   25\\ (max par for Neg) &  null& 2   &  6 \\
   \midrule
   41 \\ (max par for Pos) &  3& null   &  null \\
   \midrule
   140 \\ (max par for Neu) &  null& null   &  7 \\
  \bottomrule
\end{tabular}
\end{table}

\begin{figure}[ht]
  \centering
  \includegraphics[width=12cm]{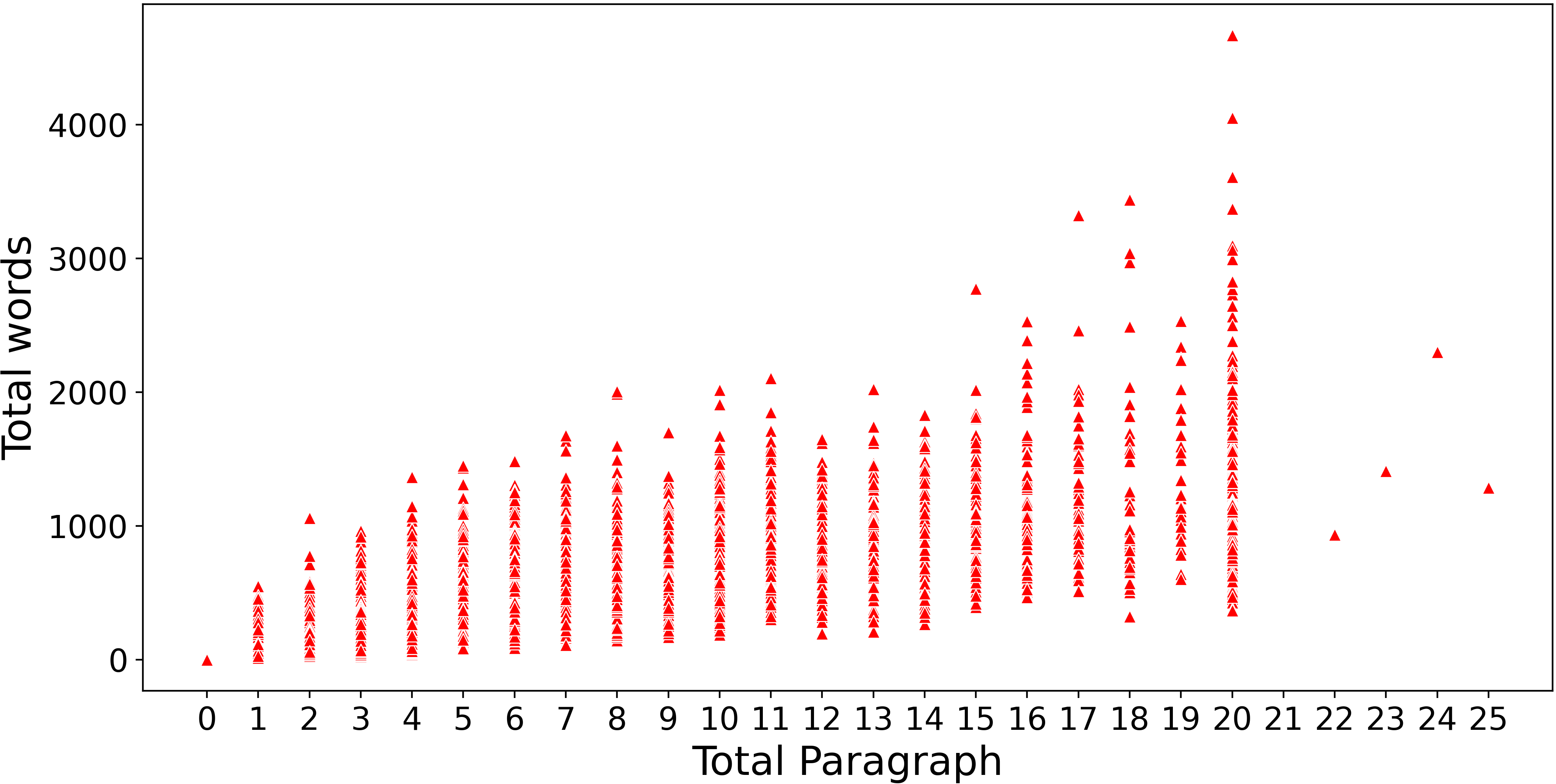}
  \caption{Relation between word counts and paragraph for negative text}
  \label{fig:words_para}
  %\Description{Relation between word counts and paragraph}
\end{figure}

The number of words that appear in each paragraph must be accountable because this defines the way how we train algorithms with respect to sequence lengths. On the other hand, it is interesting to investigate which paragraphs contribute more towards the type of labels. In other words, applicants can specify advantageous effects in the beginning paragraphs,  in middle paragraphs, or else maybe towards the end paragraphs in a list of paragraphs under one positive tag. This applies to all tags and their relevant contents. Therefore, training algorithms both at the passage level and sentence level are preferred. The example distributions of words over different paragraphs in the dataset are shown in Figure \ref{fig:words_para}.

Based on the investigation made on the dataset using data analysis, we found various issues associated such as i) null values under the tags because of format issues in XML files, ii) various duplicates, where the same content appearing in patent grants with different document numbers, and iii) data samples where word counts are less than 10 approximately.

\subsubsection{Preprocessing}

In order to mitigate the above issues, we adopted various pre-processing techniques. There are possible cases in patent drafting, such as immediately below the special tags instead of text paragraphs, there might be sub-headings that are not considered by our parsers. The usage of sub-headings, other sections or any images, etc under the special tags is dynamic and poses challenges in writing a universal parser. Furthermore, there are possible cases where authors quickly make references under special tags, this results in samples where word count less than 10 and also 0 paragraphs because of empty xml tags. Such samples with less count and other with null values are eliminated.  

Another main issue is duplicate values, for instance, USPTO list both US10516895B2 and US9866861B2 with the same title, mostly similar abstracts, also the data under special tags are completely the same. Such samples are also eliminated to avoid duplicates in the dataset. The total number of duplicates in each class of dataset is mentioned in Table \ref{tab:dataset-size}. In addition, the removal of stopwords, special tags, numbers, and other non-text matters is removed from the dataset. Keras-based\footnote{\url{https://keras.io/}} tokenization is employed to faster the execution of tokenization, where 60 seconds time is accounted for the same using Google Colab\footnote{\url{https://colab.research.google.com/notebooks/welcome.ipynb?hl=de}}. Special punctuations and capitals are avoided in order to decrease the unique words space. More detailed analysis and other exploratory experiments are available in the link\footnote{\url{https://github.com/Renuk9390/Patent-annotation/tree/main/}}.

\subsubsection{Statistics}
The total numbers of samples in dataset is around 150k (when balanced class-wise), where the distributions in each class along with duplicates are given in Table \ref{tab:dataset-size}. The average sequence length of samples ranges from 120 to 593 including all classes, on the other hand, maximum sequence length ranges from nearly 4000 to 7000 words. More about minimum sequence lengths, standard deviation, and other sequence length distributions are mentioned in Table \ref{tab:keras_tokens_distr}.  From the sequence length, it is clear that descriptions provided in the patents as solutions to problems are often lengthier, whereas advantageous effects are often shorter. It is also observed from the number of paragraphs in each class that, the maximum number of patents have their descriptions mentioned in 20 paragraphs as mentioned in Figure \ref{fig:words_para}. 

\begin{table}[ht]
\centering
  \caption{Size of dataset}
  \label{tab:dataset-size}
  \begin{tabular}{ccl}
    \toprule
  & Type&Count\\
    \midrule
     &Pos  &53475 \\

  Before pre-process  &Neg & 89105\\

  & Neu  &63055 \\
  & Total  &205635 \\
   \midrule
    
  &Pos  &48202 \\

  After pre-process  &Neg & 79531\\

  & Neu  &58043 \\
   
    & Total  &185776 \\
    \midrule
     &Pos  & 5242\\

  Duplicate values  &Neg & 9499\\

  & Neu  &4915 \\
  
    & Total  &19656 \\

  \bottomrule
\end{tabular}
\end{table}

\begin{table}[ht]
\centering
  \caption{Generated labels from 2010-2020}
  \label{tab:uspto1020}
  \begin{tabular}{ccccc}
    \toprule
     Year& 
    Total grants & 
   Positive labels & 
   Negative labels &
    Neutral labels \\
    \midrule
    2020 & 390572 & 8959 & 15307 & 11026\\
    2019 & 392618 & 9131 & 14900 & 10950\\
    2018 & 341104 & 7577 & 12080 & 9016\\
    2017 & 352547 & 7507 & 11822 & 8794\\
    2016 & 334674 & 6893 & 10546 & 7989\\
    2015 & 326969 & 6091 & 9232 & 6868\\
    2014 & 327014 & 4517 & 7211 & 5100\\
    2013 & 303642 & 2132 & 4379 & 2433\\
    2012 & 277285 & 535 & 2145 & 663\\
    2011 & 248101 & 90 & 971 & 110\\
    2010 & 244599 & 11 & 475 & 19\\
  \bottomrule
\end{tabular}
\end{table}

Table \ref{tab:uspto1020} provides a clear evidence to depict diminishing explicit subject matters as we travel back in time from 2020 to 2010. Identifying technical subject matters for imperative arguments during patent analysis manually adds a greater challenge. The top tri-grams after stopwords removal found in each class and respective counts are recorded in Table \ref{tab:trigrams}. It is observed that words such as according, invention, present are widely used phrases in the patent literature, although they do not contribute much in distinguishing classes using statistical language processing such as term frequencies and inverse document frequencies (tf-idf), their contextual relationships with other words in paragraphs may contribute. Therefore, we retain those phrases also for training algorithms.

\begin{table}[ht]
\centering
  \caption{Top tri-grams in the dataset}
  \label{tab:trigrams}
  \begin{tabular}{ccl}
    \toprule
   Class&Tri-gram&Count\\
    \midrule
  Pos   & “according present invention” &25130 \\
    & “present invention possible” &8338 \\
    \midrule
    Neg & “object present invention”& 52497\\
     & “present invention provide”& 50776\\
     \midrule
   Neu  & “aspect present invention” &51494 \\
     & “according present invention” &34298 \\
  
  \bottomrule
\end{tabular}
\end{table}

\begin{table}[ht]
\centering
  \caption{Tokens distributions using Keras  }
  \label{tab:keras_tokens_distr}
  \begin{tabular}{cccl}
    \toprule
   \textbf{Type}& & \textbf{(Class-wise counts)} &\\
   &  Pos & Neg & Neu\\
    \midrule
    mean & 120.93 &  198.10 &  593.47 \\
    
    min&  0.00& 6.00 & 2.00. \\
    25\%&  35.00&  75.00 &  193.00 \\
    50\%&  61.00&  138.00 &  410.00 \\
    75\%&  120.00&  244.00 &  819.00 \\
    std& 201.55 & 207.96  &  564.33 \\
    max&  3929.00&  4781.00 &  7317.00 \\
    
  \bottomrule
\end{tabular}
\end{table}

\section{Outcomes}
\label{sec:outcomes}
\subsection{Baseline Models}
A variety of ML models are utilized to test the performance on the dataset, where elemental features include uni-grams, bi-grams, tf-idf scores, and also NBSVM \cite{Wang12} architecture-based model. In this setting, classical ML models such as Random Forest Classifier (RFC) with maximum estimators of 200 at the maximum depth level of 3, Linear SVC (LSVC), Multinomial Naive Bayes (MNB), and Logistic Regression (LR) with random state 0, modeled with tf-idf sparse matrix of size (150000, 454182). To reckon the competence of models, we have introduced cross-validation with 5-fold, where it also discourages overfitting. The models are trained on 80\% of the total data and remaining 20\% unseen data is utilized for testing. 

Another extended model NBSVM, where core features considered are log-count ratios, technically a combination of vanilla Naive Bayes and Support Vector Machine model. NBSVM specific sequence lengths on training and test sets are recorded in Table \ref{tab:train-test-sequences}.

\begin{table}
\centering
  \caption{Train-Test Sequences}
  \label{tab:train-test-sequences}
  \begin{tabular}{ccl}
    \toprule
  & Type&Count\\
    \midrule
     &Uni-grams  &82412 \\

  Train  &Mean-uni & 306\\

  & Bi-grams  & 1990425 \\
  & Mean-bi  &612 \\
   \midrule
    
  &Mean-uni  &307 \\

  Test  &Mean-bi & 599\\

  \bottomrule
\end{tabular}
\end{table}

\subsection{Results}

The precision, recall, and F1 scores for NBSVM model are presented in Table \ref{tab:nbsvm-scores}. Considering individual class the highest F1 scores achieved by model is 97$\%$ for class $2$.   
\begin{table}[ht]
\centering
  \caption{Test scores for NBSVM}
  \label{tab:nbsvm-scores}
  \begin{tabular}{ccccc}
    \toprule
     Class& 
    Precision & 
   Recall & 
   F1-score &
    Support \\
    \midrule
    0 & 0.96 & 0.96 & 0.96 & 10028\\
    1 & 0.95 & 0.96 & 0.95 & 9930\\
    2 & 0.97 & 0.96 & 0.97 & 10042\\
  \bottomrule
\end{tabular}
\end{table}

\begin{figure}
  \centering
  \includegraphics[width=9cm]{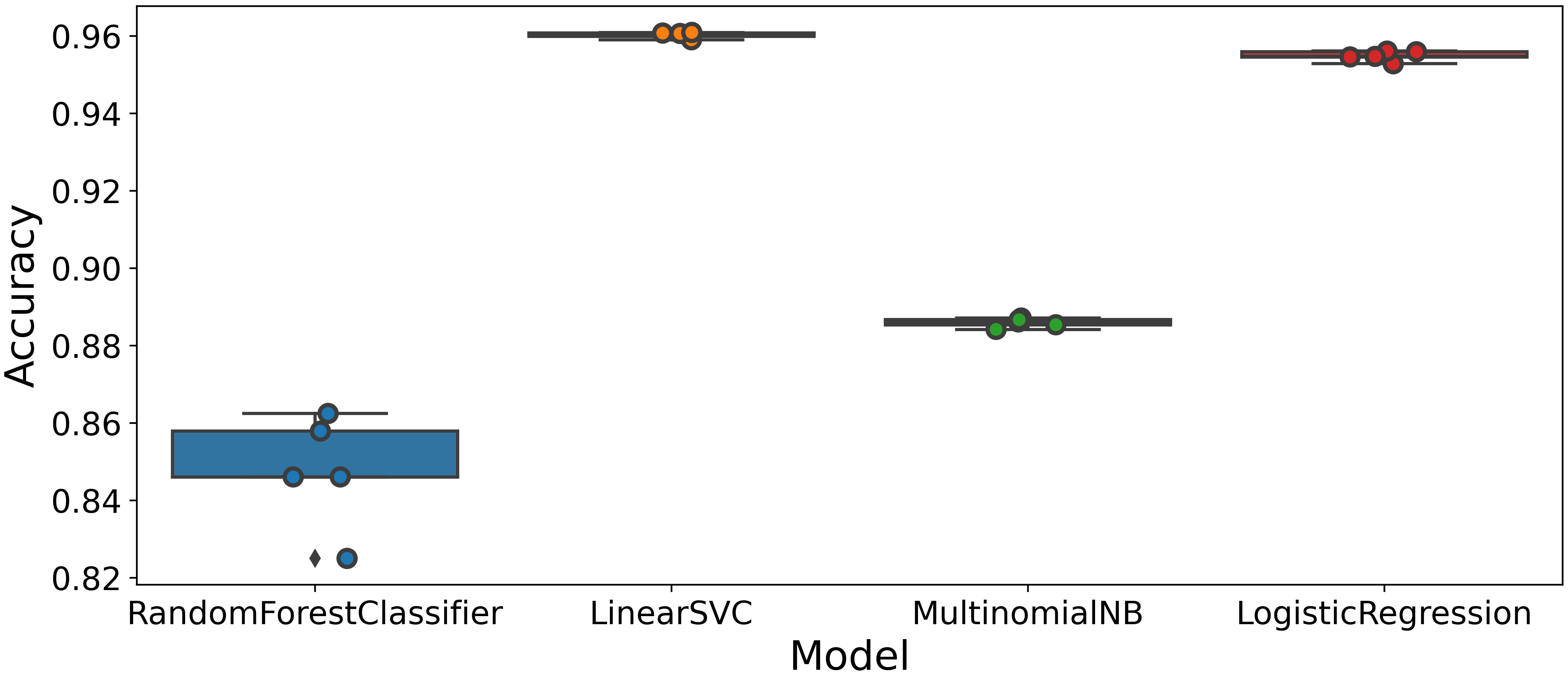}
  \caption{Accuracy scores for different models with 5 fold }
  \label{fig:k_fold_acc}
  %\Description{Accuracy scores for different models with 5 fold}
\end{figure}

\begin{figure}
% width=0.5\linewidth
  \centering
  \includegraphics[width=4cm]{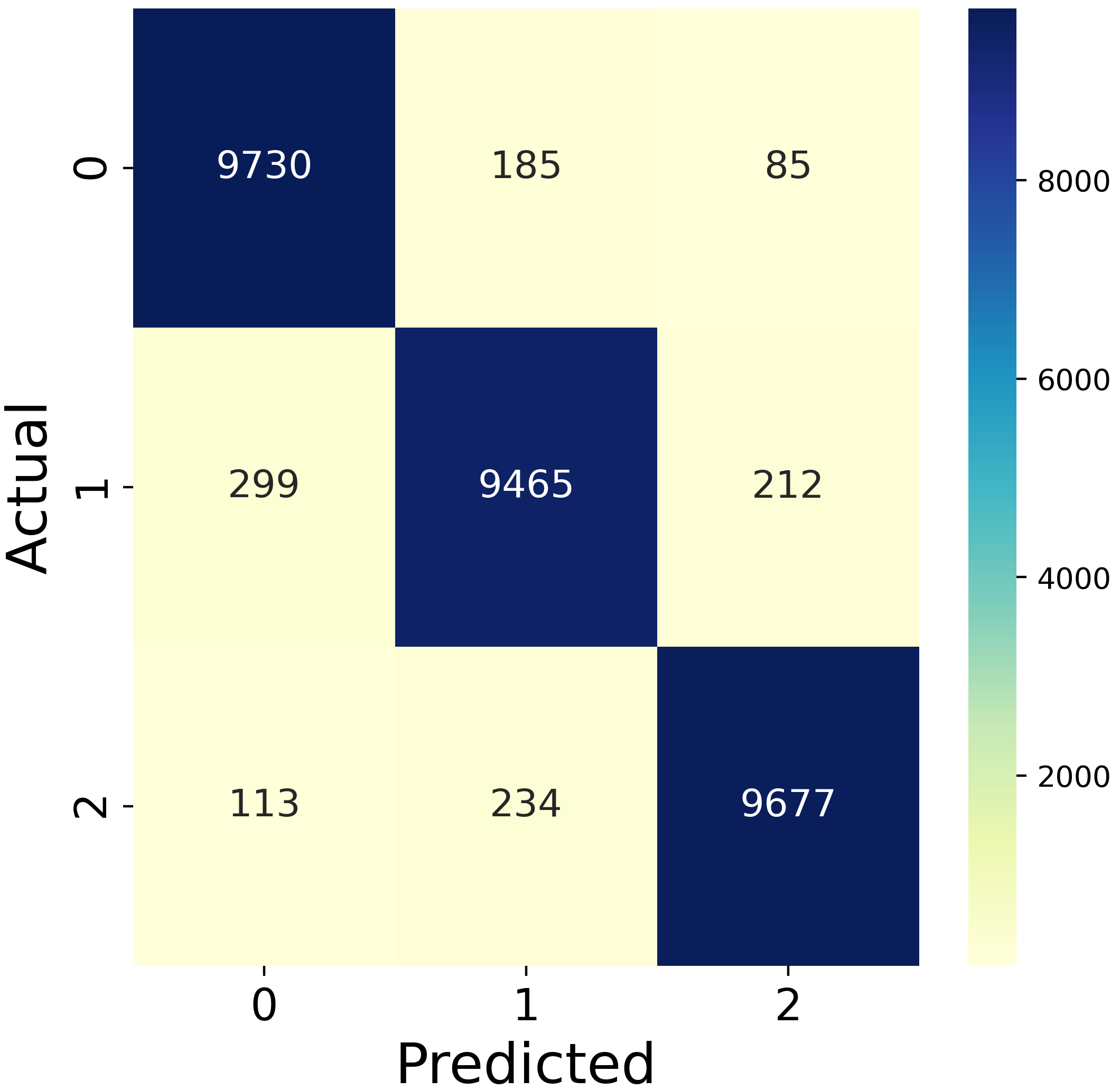}
  \caption{Confusion matrix for LSVC }
  \label{fig:cf_lsvc}
  %\Description{Confusion matrix for LSVC}
\end{figure}

The average scores over 5-fold cross validation for other ML models are recorded in Table \ref{tab:five-fold-scores}. We are considering five different ML models and noting their precision, recall and F1 scores accordingly. Considering these scores the model which performs better is the LSVC achieving F1 score of $96\%$. For the further comparison considering individual fold, box-plot is presented in Figure \ref{fig:k_fold_acc}. It shows that for the RF, its accuracy on the validation dataset varies much more for the individual folds compared to other three models. To get insight of a model performance the confusion matrix for the LSVC is shown in Figure \ref{fig:cf_lsvc} for all three labels. The quality of the dataset determines the way how algorithms are trained, and further influences the performances. The labels considered in this work are reliable, consistent, and ground truth evidence is taken from the patent text such that explicitly mentioned tags by applicants are sourced. We anticipate that our attempt helps in identifying subjective information in patent documents automatically.  Introducing a new dataset for patent sentiment analysis and adding a quick usage of this data using ML models are being the focus of this work, therefore extended experiments and complex DL architectures can be seen in the future work.  

\begin{figure*}[ht]
  \centering
  \includegraphics[width=15cm]{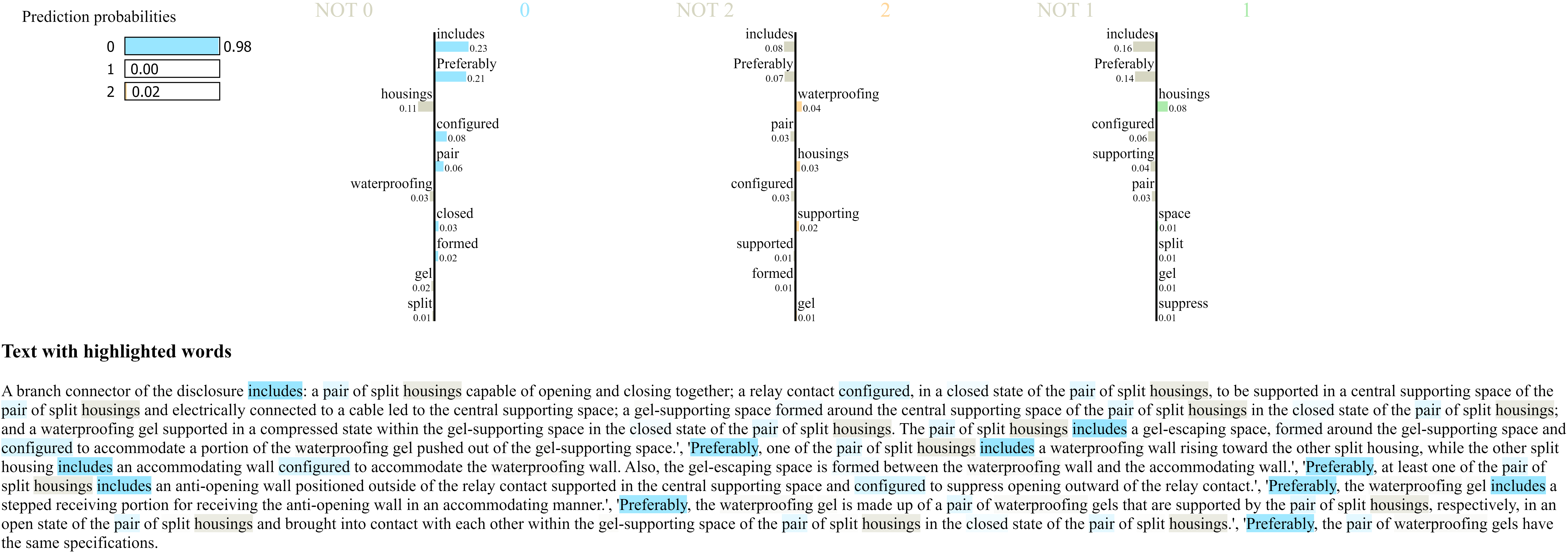}
  \caption{Visualising the LSVC model prediction for the text and deciding neutral label}
  \label{fig:lime_LSVC}
  %\Description{Visualising the LSVC model prediction for the text and deciding neutral label}
\end{figure*}

\begin{figure*}[ht]
  \centering
  \includegraphics[width=15cm]{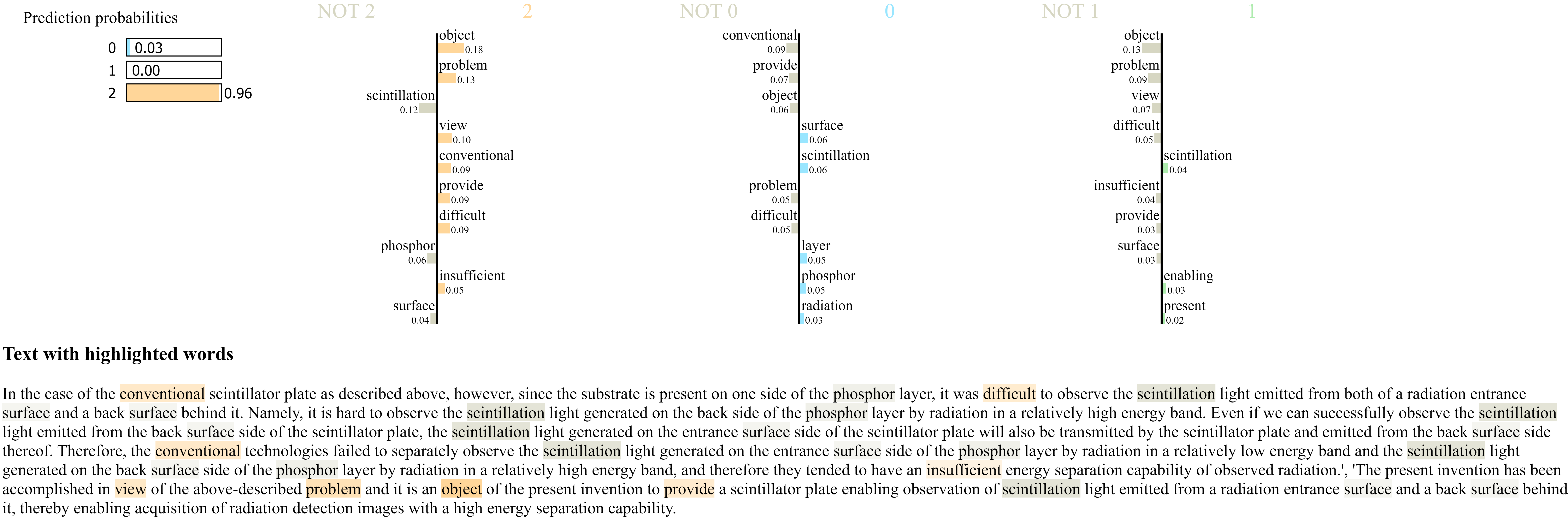}
  \caption{Visualising the LSVC model prediction for the text and deciding negative label}
  \label{fig:lime_LSVC_neg}
 % \Description{Visualising the LSVC model prediction for the text and deciding negative label}
\end{figure*}

To visualize how specific words are contributing to deciding appropriate labels using the trained model, Python package Lime\footnote{\url{https://github.com/marcotcr/lime}} is integrated. Figure \ref{fig:lime_LSVC} and \ref{fig:lime_LSVC_neg} shows that how the label $0$ and $2$ are assigned to the text. It shows top $10$ words with their probability in deciding whether they are in accordance with a label (i.e. $0$, $1$, $2$) or not (i.e. NOT $0$, NOT $1$, NOT $2$). Furthermore, it is highlighted in the text with the specific word and their respective colour code for better insight.   

\begin{table}[ht]
\centering
  \caption{Average 5-fold scores}
  \label{tab:five-fold-scores}
  \begin{tabular}{cccc}
    \toprule
     Model& 
    Precision & 
   Recall & 
    F1-score  \\
    \midrule
    RFC & 0.84 & 0.85 & 0.85 \\
    LSVC & 0.96 & 0.96 & 0.96 \\
    MNB & 0.89 & 0.89 & 0.89 \\
    LR & 0.95 & 0.96 & 0.95 \\
  \bottomrule
\end{tabular}
\end{table}

\section{Conclusion and Future Work}
\label{sec:conclusion}
Patent paragraph highlighting is a form of patent sentiment analysis and information retrieval. It is considered as a crucial stage in the assessment of patent documents. Patent practitioners such as examiners and attorneys have a practice of aligning their arguments found for technical subject matters of any invention, such alignments are effective through bringing automation in highlighting technical entities which are necessary for comparing inventions to existing literature. To this end, we propose a novel dataset to train ML algorithms to highlight patent paragraphs automatically based on individual subject matter types. The USPTO full-text grants (for over a decade) in the form of XML files are considered to develop the dataset. Initial baseline models using ML to utilize the dataset and their results are added, which helps other researchers to extend work for further refinements. The source codes to collect the raw data, parse the data to retrieve informative text segments, exploratory data analysis, and model-related codes along with links to download dataset are provided to reinforce patent information retrieval. 

Future work to this end includes: i) enhance the performances of baseline models using DL with complex learning capabilities, ii) introducing domain-specific pre-trained language models to enforce highlighting tasks, and iii) deployment of extended models as an API to build patent paragraph highlighting tool as a browser extension. Other enhancements towards the dataset are also possible and interesting future work such as i) extending the dataset for various other possible tags, ii) focusing on other patent office grants to make the dataset more universal.  We assert that, this work stands as preliminary and opens a new path to bring automation in highlighting tasks in order to ease examination process and also to aid other patent practices. 

\bibliographystyle{unsrtnat}
\bibliography{references}  %%% Uncomment this line and comment out the ``thebibliography'' section below to use the external .bib file (using bibtex) .





% \end{thebibliography}

\end{document}